\documentclass[conference]{IEEEtran}
\IEEEoverridecommandlockouts
\usepackage{cite}
\usepackage{amsmath,amssymb,amsfonts}
\usepackage{algorithmic}
\usepackage{graphicx}
\usepackage{textcomp}
\usepackage{xcolor}
\usepackage{orcidlink} 
\usepackage{caption}

\usepackage{hyperref} 
\usepackage{float}

\usepackage{array,booktabs}
\usepackage{multirow}
\usepackage[inline]{enumitem}

\def\BibTeX{{\rm B\kern-.05em{\sc i\kern-.025em b}\kern-.08em
    T\kern-.1667em\lower.7ex\hbox{E}\kern-.125emX}}
\begin{document}

\title{Foundation CAN LM: A Pretrained Language Model For Automotive CAN Data\\
}

\author{
    \IEEEauthorblockN{
        Akiharu Esashi\textsuperscript{1,3}\orcidlink{0009-0001-6802-336X},
        Pawissanutt Lertpongrujikorn\textsuperscript{1}\orcidlink{0009-0003-4106-2347},
        Justin Makino\textsuperscript{2},
        Yuibi Fujimoto\textsuperscript{3},
        Mohsen Amini Salehi\textsuperscript{1}\orcidlink{0000-0002-7020-3810}
    }

    \IEEEauthorblockA{\textsuperscript{1}\href{https://hpcclab.org}{HPCC Lab}, University of North Texas, Denton, Texas, USA\\
    \texttt{\{akiharu.esashi, pawissanutt.lertpongrujikorn, mohsen.aminisalehi\}@unt.edu}}

    \IEEEauthorblockA{\textsuperscript{2}Connected Analytic Services, Plano, Texas, USA\\
    \texttt{justin.makino@connectedanalyticservices.com}}

    \IEEEauthorblockA{\textsuperscript{3}Toyota Insurance Management Solutions, Plano, Texas, USA\\
    \texttt{yuibi.fujimoto@toyota-ims.com}}
}

\maketitle
\begingroup
\renewcommand\thefootnote{}
\footnotetext{Accepted Preprint at the 37th IEEE Intelligent Vehicles Symposium (IV 2026).}
\endgroup

\begin{abstract}
The Controller Area Network (CAN) bus provides a rich source of vehicular signals increasingly leveraged for applications in automotive and auto insurance domains, including collision detection, predictive maintenance, and driver risk modeling. Despite this potential, existing pipelines largely train isolated task-specific models on raw CAN data, with only limited efforts exploring decoded signals. Such fragmentation prevents shared representation learning and limits cross-task generalization. By contrast, natural language processing (NLP) and computer vision (CV) have been transformed by the foundation model paradigm: large-scale pretraining followed by task-specific adaptation. In this work, we introduce the \emph{foundation CAN model} that demonstrates multi-objective downstream generalization using a single pretrained backbone. Our approach treats CAN data as a language: we pretrain on large-scale, unlabeled decoded CAN signals and fine-tune across heterogeneous auto insurance tasks. To enable this, we propose a unified tokenization scheme for mixed discrete--continuous signals and address challenges of temporal complexity and trip-specific variability. Our results show that one pretrained CAN model can adapt effectively to diverse predictive tasks, validating that the foundation modeling paradigm, proven in NLP and CV, also holds for CAN data. This establishes a new direction for generalizable representation learning in automotive AI.

\end{abstract}

\begin{figure*}[t] \centering 
\vspace{-5pt}

\includegraphics[width=0.95\linewidth]{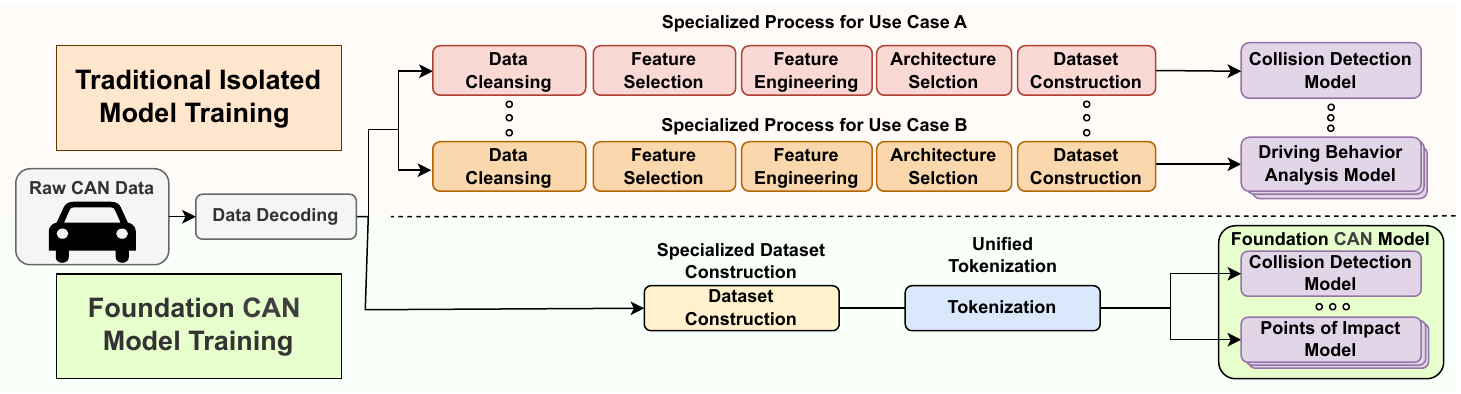} 

\caption{\normalsize Traditional task-specific pipelines (\textbf{top}) require isolated preprocessing and models for each task. Our foundation model approach (\textbf{bottom}) uses unified tokenization and a single pretrained backbone that adapts to multiple downstream tasks through task-specific fine-tuning.} \label{fig:architecture} 
\vspace{-10pt} 
\end{figure*}

\section{Introduction}

The Controller Area Network (CAN) bus \cite{szydlowski1992can} is the backbone of in-vehicle communication, enabling Electronic Control Units (ECUs) to exchange information such as speed, braking signals, and engine status \cite{oguma2008new}. With proper decoding, CAN data exposes rich signals that are increasingly leveraged by both the automotive and auto insurance industries. Applications span collision detection and avoidance, predictive maintenance, and autonomous driving on the automotive side, as well as driver behavior scoring, point-of-impact detection, and total loss assessment in the auto insurance domain \cite{fugiglando2018driving, hanselmann2020canet, abbache2025toward }.  
Despite this growing interest, current AI/ML applications of CAN data remain fragmented. Most existing pipelines train isolated, task-specific models directly on raw CAN traffic, with a few recent efforts exploring decoded signals \cite{canbert-correct, nam2021intrusion, nguyen2023tan, forklift2024}. This fragmented approach prevents shared representation learning, incurs redundant data preparation and training costs, and limits generalization across tasks (Fig.~\ref{fig:architecture}, top). In contrast, fields such as natural language processing (NLP) and computer vision (CV) have been transformed by the emergence of \textit{foundation models}\cite{myers2024foundation, du2022survey}: large-scale pre-training followed by task-specific post-training, yielding general-purpose backbones that can be fine-tuned efficiently for diverse downstream applications \cite{qiu2024transferring}. Yet, no comparable framework has demonstrated such generalization for CAN data.  

We identify this gap as the central research problem: while prior work has shown that CAN traffic can be modeled with transformer-based pretraining \cite{devlin2019bert} for a \emph{single objective}, there is no evidence of a generalizable pretraining paradigm for CAN signals that transfers effectively across \emph{multiple} automotive tasks. To address this, we propose treating CAN data as a language and applying the Large Language Model (LLM) paradigm (Fig.~\ref{fig:architecture}, bottom). In this approach, a single foundation model is pretrained on large-scale, unlabeled decoded CAN data and subsequently adapted via fine-tuning to heterogeneous downstream objectives. This enables shared representation learning, reduces redundancy, and improves cross-task generalization, overcoming the limitations of the current isolated-model paradigm.  

Developing a foundation CAN model, however, presents unique challenges beyond those faced in text or vision domains. First, tokenization is non-trivial: unlike text with discrete symbols, CAN signals contain mixed discrete-continuous values, requiring a robust, reproducible scheme. Second, CAN data is inherently a multi-scale time series, with dependencies ranging from millisecond-level sensor dynamics to long-horizon trip-level patterns. Third, each trip introduces unique contextual factors (vehicle state, driver, environment), demanding representations that generalize across trips while capturing within-trip dynamics.  

In this work, we make the following contributions:
\begin{enumerate}
    \item \textbf{Multi-Objective Foundation CAN Model}: We introduce the pretraining--post-training framework that demonstrates cross-task generalization from a single CAN backbone, enabling diverse automotive tasks to benefit from shared representations.  
    \item \textbf{Unified Tokenization for Mixed CAN Signals}: We design a reproducible tokenization scheme tailored to CAN data, integrating scaling and quantization to handle both discrete and continuous signals for large-scale pretraining.  
    \item \textbf{Cross-Task Evaluation Study}: We provide a systematic study of a pretrained CAN model across heterogeneous downstream tasks, validating that the foundation modeling paradigm, proven in NLP and CV, also holds for CAN data.  
\end{enumerate}

\section{Related Work}


\noindent\textbf{Task-specific modeling of CAN signals}: Most learning pipelines over CAN data train isolated models for a single objective; e.g., intrusion detection system (IDS) \cite{ hoang2023canperfl}, driver identification \cite{kim2024timeshift}, or predictive maintenance \cite{abbache2025toward}, directly from raw CAN traffic. While effective within narrow domains, this fragmentation prevents shared representation learning, duplicates training effort across tasks, and limits generalization beyond the target application.


\noindent\textbf{Transformer-based modeling of raw CAN traffic}: Several works adapt Transformer architectures and language-modeling objectives to the in-vehicle setting. \textsc{CAN-BERT}~\cite{canbert-correct} employs masked language modeling (MLM) over raw CAN messages for IDS, and Nam~\emph{et~al.}~\cite{nam2021intrusion} propose a bi-directional GPT-style model for the same task. Nguyen~\emph{et~al.}~\cite{nguyen2023tan} introduce an Attention Network that improves IDS efficiency and shows partial cross-vehicle transfer. While these works validate that raw CAN streams can be modeled with transformers, they remain confined to \emph{single-task} objectives, and do not explore decoded signals or multi-task generalization.

\noindent\textbf{Decoded CAN signal modeling}: Beyond raw CAN data, a few recent efforts explore decoded CAN signals \cite{forklift2024, fugiglando2018driving}. The thesis \emph{CAN you speak Forklift?}~\cite{forklift2024} demonstrates GPT-style pretraining on decoded industrial-vehicle signals for activity recognition. This work tokenizes forklift CAN signals by extracting state-transition events from \emph{four signals} and models them autoregressively to predict operational sequences. However, the approach remains domain-specific (industrial forklifts vs. consumer vehicles), relies on event-based preprocessing that discards intra-event dynamics, operates on a limited signal set (4 vs. our 44 decoded features), and critically, does not evaluate cross-task transfer. Our work addresses these limitations by pretraining on a comprehensive set of decoded CAN signals with unified tokenization and demonstrating adaptation across heterogeneous tasks.

\noindent\textbf{Research gap and positioning}: The literature demonstrates that (i) transformers can model CAN sequences effectively, (ii) decoded signals enable richer semantic modeling than raw traffic, and (iii) foundation-style pretraining benefits adjacent automotive domains. Yet, to the best of our knowledge, \emph{no prior work has demonstrated that a single pretrained CAN backbone can adapt across multiple heterogeneous automotive tasks}. Existing CAN models target single objectives, and the limited decoded-signal work does not establish cross-task transfer. Our work bridges this gap by (i) pretraining an LM with MLM over a unified mixed discrete--continuous vocabulary spanning 44 decoded automotive sensor signals, and (ii) demonstrating adaptation to heterogeneous downstream tasks. This transition from task-specific modeling to a CAN \emph{foundation model} establishes the viability of multi-objective downstream generalization for automotive and auto insurance applications (aka downstream tasks).

\section{Methodology}

Our methodology follows a two-stage paradigm: (1) large-scale pretraining on unlabeled decoded CAN signals using MLM, and (2) task-specific fine-tuning for heterogeneous downstream objectives. Central to this approach is a unified tokenization framework that bridges the gap between discrete linguistic representations and mixed discrete--continuous CAN signals. We first describe the dataset and preprocessing (\S\ref{sec:data}), then detail our tokenization strategy (\S\ref{sec:tokenization}), and the pretraining procedure (\S\ref{sec:pretraining}).

\subsection{Dataset and Preprocessing}
\label{sec:data}

The pretraining dataset comprises fully anonymized, non-identifiable CAN data, decoded from ${\sim}10{,}000$ vehicles over a 90-day period from a leading global automotive OEM, with a subset of 9 days ($\sim$19B tokens) used for initial training. Each trip log is decoded into 44 interpretable features in four domains: (i) vehicle dynamics (e.g., speed, acceleration), (ii) driver behavior (e.g., brake, gear), (iii) safety indicators (e.g., collision warning), and (iv) vehicle state and context (e.g., doors, occupancy). Decoded CAN data is synchronized to a 1~Hz frequency, normalized within empirically determined operating ranges, and segmented into 10-second windows, producing 450 tokens per sequence after tokenization.


\begin{figure}[t]
\centering
\resizebox{1.0\linewidth}{!}{%
    \includegraphics{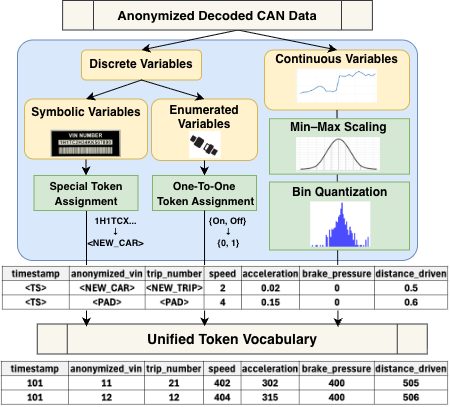}
}
\vspace{-10pt}
\caption{\normalsize Unified tokenization pipeline for mixed-mode CAN signals. Decoded CAN data is split into discrete variables (symbolic and enumerated) and continuous variables, processed through type-specific strategies, then merged into a unified token vocabulary for pretraining.}
\label{fig:4process}
\vspace{-10pt}

\end{figure}



\subsection{Unified Tokenization of Mixed Signals}

\label{sec:tokenization}

Unlike natural language, where words belong to a discrete vocabulary, CAN signals are composed of mixed discrete and continuous variables. Each decoded CAN frame consists of three signal types:
\begin{enumerate*}
    \item \textbf{Continuous variables} (e.g., speed, steering angle): numeric, real-valued, and smoothly varying over time.
    \item \textbf{Discrete variables} (e.g., gear, door status): finite, enumerated categorical values.
    \item \textbf{Symbolic identifiers} (e.g., anonymized vehicle identification number (VIN), trip ID): identifiers representing context boundaries between trips and vehicles.
\end{enumerate*}

Tokenization is therefore non-trivial and central to enabling foundation-style pretraining. A naïve one-to-one mapping of continuous values would produce an unbounded, non-reproducible vocabulary. To address this, we develop a \emph{unified tokenization framework} for mixed-mode CAN signals that ensures semantic fidelity, reproducibility, and transferability across datasets (Fig.~\ref{fig:4process}). This framework employs a fixed, predefined binning and enumeration schema to provide interpretable mappings between feature values and token IDs, emphasizing reproducibility and cross-dataset consistency.

\vspace{1mm}
\noindent\textbf{Continuous-Variable Tokenization.}
For continuous signals, we employ an empirically calibrated, feature-specific quantization strategy to ensure reproducible and scale-aware tokenization:
\begin{enumerate}[noitemsep, leftmargin=*]
    \item \textbf{Outlier Handling:} Each feature’s value is compared against pre-defined minimum and maximum thresholds obtained from the static sensor range limits. Values outside these bounds are replaced with a dedicated \texttt{<OUTLIER>} token, predefined invalid entries are replaced with \texttt{<ERROR>}, and \texttt{<NULL>} for missing values. Separating these cases enables the model to differentiate between fundamentally different failure modes and operational situations.
    
    \item \textbf{Normalization:} Valid values are linearly scaled to the range [0,1] using 
\textit{empirically determined} min–max bounds specific to each feature.
Unlike OEM-specified sensor ranges, which are often significantly broader 
than values observed in real driving, these empirical limits are estimated from nine days of CAN data collected across ${\sim}10{,}000$ vehicles. 
This captures realistic operating intervals while excluding synthetic or rare extreme values that distort scaling. 
Such calibration reduces token sparsity by preventing overly fine quantization 
of unoccupied value regions and ensures reproducible normalization 
even when OEM scaling or sensor characteristics differ across datasets.

    \item \textbf{Discretization:} 
Each continuous feature is discretized into a fixed number of 
uniform bins (e.g.,128, 256) assigned from a preconfigured 
mapping based on its normalized temporal variation $r_i$ 
(Eq.~\ref{eq:discretization_metric}). 
To estimate $\Delta_i$, approximately 100 vehicles × 15 trips were sampled; 
each sequence was ordered by timestamp, clipped to valid empirical ranges, 
and the upper and lower 0.5\% of $|x_{t+1} - x_t|$ values were excluded 
before averaging to suppress transient spikes. 
Features with smaller $r_i$ receive finer bins, while more dynamic signals use coarser ones based on preset thresholds, 
ensuring reproducible yet scale-aware tokenization across datasets.

\vspace{-15pt}

\begin{equation}
r_i = \frac{\Delta_i}{\max(x_i) - \min(x_i)}, 
\quad 
\Delta_i = \mathbb{E}\left[\,|x_{t+1} - x_t|\,\right],
\label{eq:discretization_metric}
\end{equation}

\end{enumerate}

This fixed-bin discretization produces reproducible token boundaries and maintains a consistent representation scheme across all datasets.

\vspace{1mm}
\noindent\textbf{Discrete-Variable Tokenization.}
Discrete variables are further divided into two subtypes:
\begin{enumerate}[noitemsep, leftmargin=*]
    \item \textbf{Enumerated states:} low-cardinality categorical variables (e.g., turn signal, cruise control = \{On, Off\}) are mapped one-to-one to categorical tokens.
    \item \textbf{Symbolic identifiers:} Context markers such as anonymized VIN and trip ID are abstracted into meta-tokens \texttt{<NEW\_CAR>} and \texttt{<NEW\_TRIP>} to denote context shifts without inflating the vocabulary. 
    Analogous to special tokens in NLP (e.g., \texttt{<BOS>} or \texttt{<SEP>})~\cite{devlin2019bert}, these meta-tokens serve as boundary markers indicating transitions between distinct contexts, such as a new vehicle instance or trip.

    During tokenization, a meta-token is inserted whenever the VIN or trip ID changes. For all subsequent samples within the same vehicle or trip context, the corresponding identifier field is filled with a \texttt{<PAD>} token, indicating that no context change has occurred.
    
\end{enumerate}

This distinction allows the model to learn intra-trip temporal dependencies while maintaining generalization across trips and vehicles.

\vspace{1mm}
\noindent\textbf{Unified Vocabulary Construction.}
All tokens derived from continuous and discrete variables are merged into a unified vocabulary of approximately 1{,}420 unique tokens, including other general special tokens such as \texttt{<MASK>} and \texttt{<CLS>}. 
In addition, infrastructure-level special tokens are included to encode sequence structure. \texttt{<TS>} (timestamp marker) is added to indicate the start of each signal sequence, thereby explicitly encoding temporal boundaries in the serialized token stream.

\noindent\textbf{Sequencing of Tokens Across Time.}
After each signal value is converted into its corresponding token, we serialize the tokens into a fixed ordering that preserves temporal structure (see the bottom part of Fig.~\ref{fig:4process}). At every timestamp (1 Hz), a \texttt{<TS>} and all 44 decoded CAN features are tokenized independently and then concatenated in a consistent, predetermined feature order to form a single time-step block of tokens. \texttt{<TS>} token is inserted at the beginning of each time-step block to mark temporal boundaries and allow the model to learn cross-feature dependencies within a moment as well as transitions across time. For a 10-second window, this produces a sequence of 10 \texttt{<TS>} markers followed by their associated feature tokens, resulting in 450 total tokens. This ordering ensures that the Transformer processes CAN data as a coherent, structured sequence analogous to a sentence, where temporal progression is expressed through repeated feature-wise token groups.

\subsection{Foundation Model Pretraining}
\label{sec:pretraining}

We pretrain a Transformer-based foundation model on large-scale, unlabeled decoded CAN signals to learn shared representations transferable across heterogeneous automotive tasks. The model is trained using a \textit{MLM} objective, treating CAN signals as structured sequences analogous to natural language.

\vspace{1mm}
\noindent\textbf{Model Architecture}
Tokenized sequences are used to train a Bidirectional Encoder Representations
from Transformers (BERT) style Transformer encoder with an MLM objective without Next Sentence Prediction (NSP). Following the original pre-training setup~\cite{devlin2019bert}, the model minimizes the cross-entropy loss between the predicted token distributions and the original masked tokens. Fifteen percent of tokens in each sequence are randomly selected for masking: 80\% are replaced with the special \texttt{<MASK>} token, 10\% with random tokens, and the remaining 10\% are left unchanged. The model learns to reconstruct masked tokens using bidirectional context, capturing dependencies across temporal and semantic dimensions.

\vspace{1mm}
\noindent\textbf{Pretraining Validation}
Model learning progress is evaluated via MLM loss convergence and internal t-SNE visualization of learned embeddings. Tokens corresponding to semantically related signals (e.g., speed and acceleration) form coherent clusters as well as moderate intra-clusters (e.g., speed 0 - speed 100), indicating successful contextual learning. Due to proprietary constraints, these visualizations are not included. The results confirm that the proposed tokenization and pretraining pipeline effectively capture shared structure across heterogeneous CAN signals.

\section{Performance Evaluation}
\label{sec:evaluation}
We evaluate the proposed foundation CAN model to verify its capacity for cross-task generalization and downstream adaptability. The evaluation focuses on how a large-scale pretrained representation, trained solely on decoded CAN data, transfers to specialized tasks with different label structures and class distributions. Importantly, the objective is not to outperform task-optimized or heavily engineered baselines on individual tasks, but to assess whether a single pretrained backbone can adapt across heterogeneous objectives. Rather than emphasizing absolute scores, we aim to demonstrate the robustness of the evaluation design in assessing the model’s adaptability to diverse objectives.

\begin{figure*}[t]
\vspace{-10pt}

\centering

\raisebox{0.6\height}{%
\begin{minipage}{0.48\textwidth}
\centering

\begin{tabular}{lcccc}

\toprule
\textbf{Ratio} & \textbf{Model} & \textbf{Precision} & \textbf{Recall} & \textbf{F1} \\
\midrule

\multirow{2}{*}{\textbf{10:1}}
 & \textbf{GLM}            & \textbf{96.9\%} & \textbf{69.8\%} & \textbf{81.1\%} \\
 & \textbf{Foundation CAN} & \textbf{94.7\%} & \textbf{63.7\%} & \textbf{76.1\%} \\
\midrule

\multirow{2}{*}{\textbf{100:1}}
 & \textbf{GLM}            & \textbf{76.8\%} & \textbf{56.8\%} & \textbf{65.3\%} \\
 & \textbf{Foundation CAN} & \textbf{60.9\%} & \textbf{48.9\%} & \textbf{54.2\%} \\
\bottomrule
\end{tabular}
\vspace{5pt}

\captionof{table}{\normalsize Evaluation results for binary classification across imbalance ratios.}
\vspace{-3pt}

\label{tab:binary_metrics}
\end{minipage}
}
\hfill
\raisebox{0.5\height}{%
\begin{minipage}{0.24\textwidth}
\centering
\includegraphics[width=\linewidth]{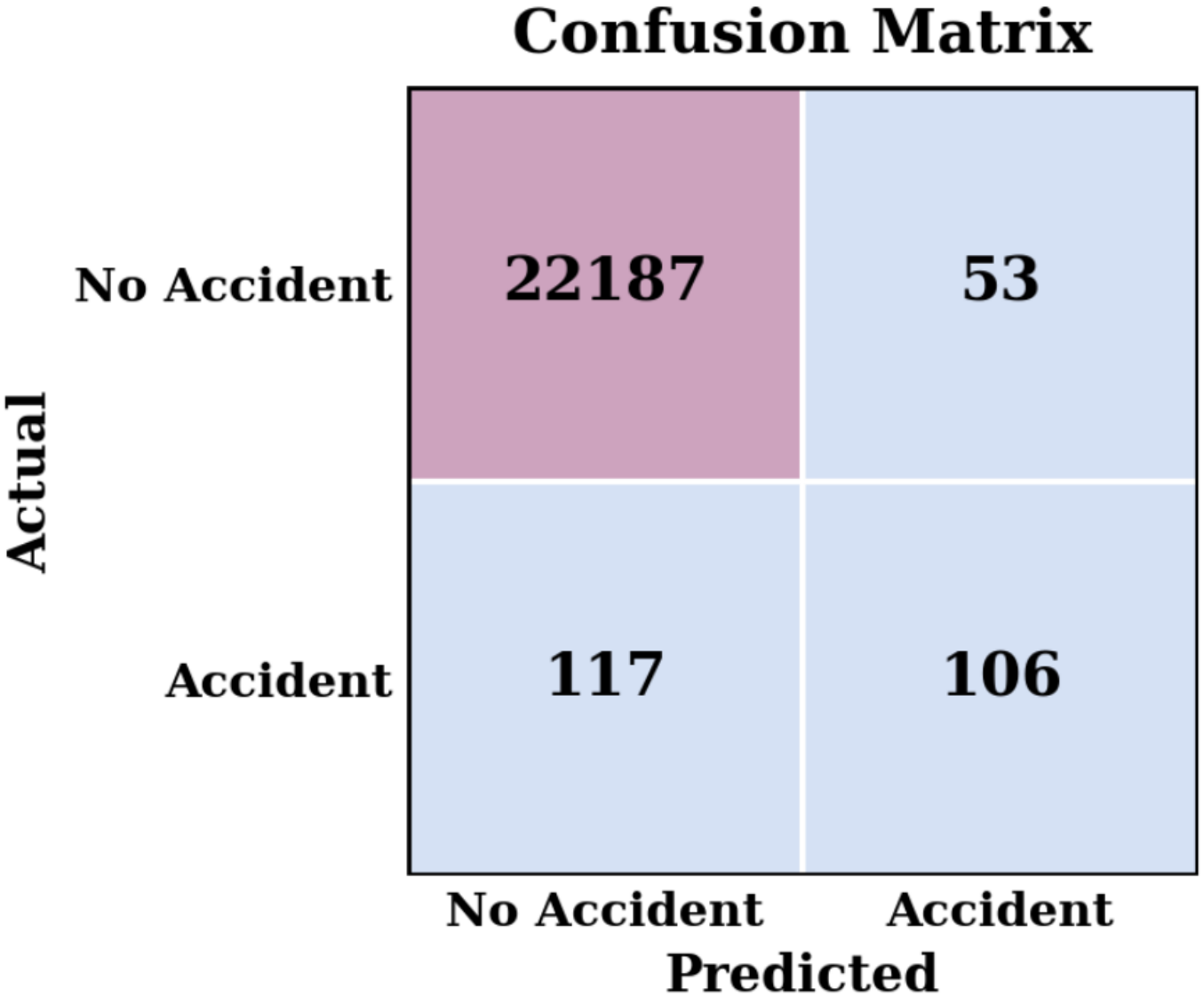}
\vspace{-15pt}
\caption{\normalsize Foundation CAN Model (10:1 Ratio)}
\vspace{8pt}

\label{fig:foundation_confusion}
\end{minipage}
}
\hfill
\raisebox{0.5\height}{%
\begin{minipage}{0.24\textwidth}
\centering
\includegraphics[width=\linewidth]{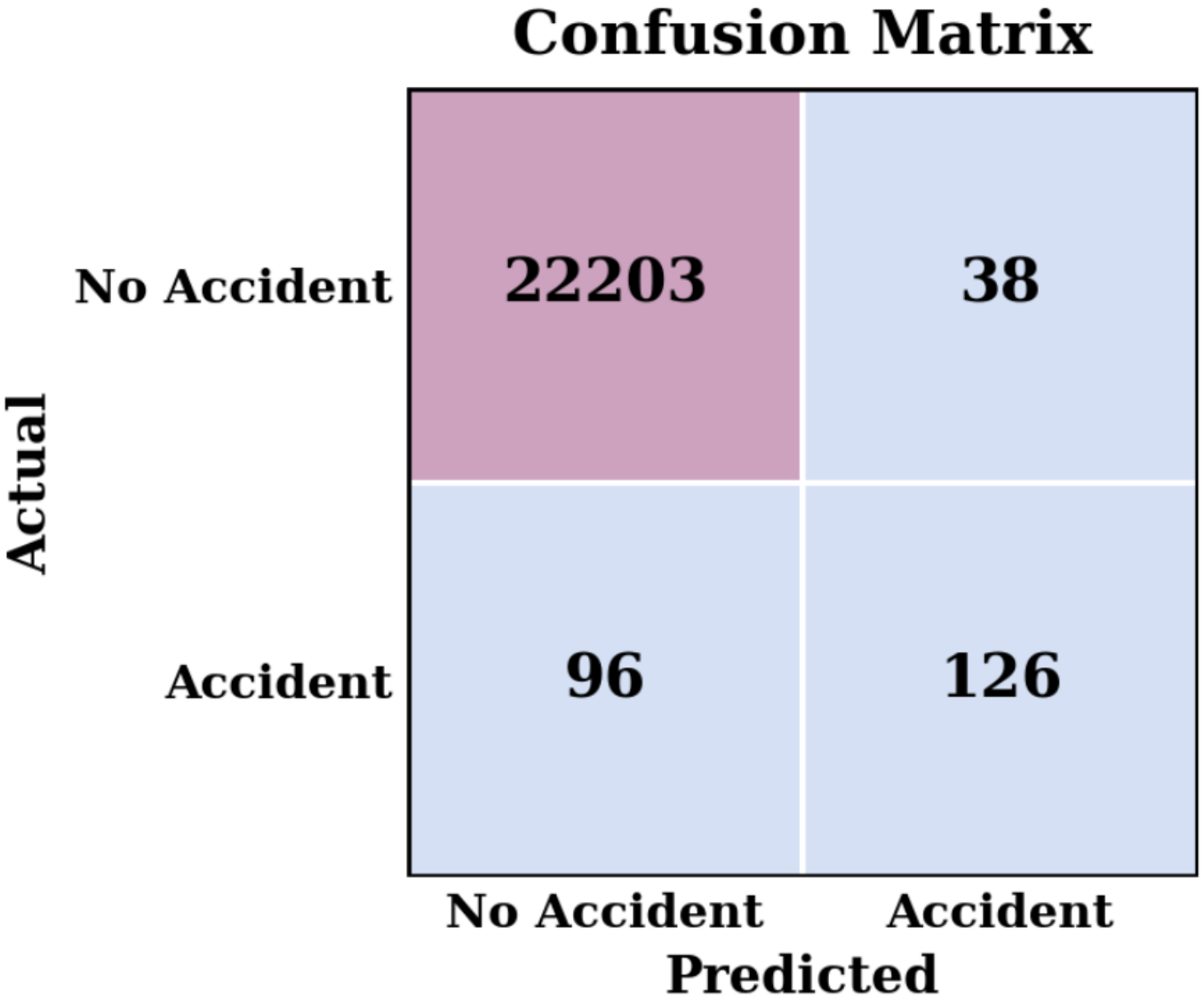}
\vspace{-15pt}

\caption{\normalsize Baseline GLM (10:1 Ratio)}
\label{fig:baseline_confusion}
\vspace{10pt}

\end{minipage}
}

\end{figure*}

\begin{table*}[t]
\centering
\vspace{-10pt}

\normalsize     

\vspace{10pt}

\begin{tabular}{lcccccc}
\toprule
\textbf{Model} &
\multicolumn{2}{c}{\textbf{Precision}} &
\multicolumn{2}{c}{\textbf{Recall}} &
\multicolumn{2}{c}{\textbf{F1}} \\
\cmidrule(lr){2-3} \cmidrule(lr){4-5} \cmidrule(lr){6-7}
& \textbf{Macro} & \textbf{Weighted}
& \textbf{Macro} & \textbf{Weighted}
& \textbf{Macro} & \textbf{Weighted} \\
\midrule
\addlinespace[5pt]

\textbf{CNN} 
& \textbf{30.4\%} & \textbf{34.6\%}
& \textbf{27.8\%} & \textbf{30.9\%}
& \textbf{24.3\%} & \textbf{27.4\%} \\

\addlinespace[5pt]

\textbf{Foundation CAN}
& \textbf{28.3\%} & \textbf{33.1\%}
& \textbf{28.2\%} & \textbf{34.9\%}
& \textbf{27.0\%} & \textbf{32.6\%} \\

\addlinespace[5pt]
\bottomrule
\end{tabular}
\vspace{-2pt}
\caption{\normalsize\normalfont Evaluation results for multiclass classification. Macro averages treat all classes equally, while weighted averages account for class imbalance by giving more weight to classes with more samples.}
\label{tab:multiclass_metrics}
\vspace{-10pt}
\end{table*}

\subsection{Experimental Setup}
Our experiments utilize a single pretrained model comprising 9 transformer encoder layers with a hidden size of 670, 10 self-attention heads, and an intermediate feed-forward dimension of 2{,}680, totaling $\sim$50M parameters, serving as a unified backbone across all downstream tasks.
We adopt full-parameter fine-tuning, updating all encoder weights alongside a newly initialized classification head, and apply class-weighted loss to address severe class imbalance.
This design isolates the transferability of pretrained representations from task-specific architectural choices, directly evaluating foundation model generalization in the automotive CAN domain.


To ensure that the evaluation reflects realistic deployment conditions, we compare against established baselines currently deployed in operational contexts.
These include Generalized Linear Models (GLM) with a logit link \cite{mccullagh2019generalized} and 1D Convolutional Neural Networks (CNN) \cite{o2015introduction}, both of which represent widely adopted and well-validated approaches in automotive and insurance applications.
These baselines provide a meaningful benchmark precisely because they incorporate domain expertise and have demonstrated effectiveness in real-world settings, making them representative of the performance standards our foundation model to compare for practical adoption.

All tasks employ decoded CAN data, signals are uniformly resampled at 1~Hz to match the pretraining configuration, with each segment 10-second sequences representing the driving history corresponding to a 450-token input sequence.
For supervised learning, collision labels are derived through temporal alignment between insurance records, GPS traces, and verified police accident reports.
The resulting dataset captures both normal driving dynamics and rare collision scenarios, providing an ideal benchmark for evaluating representation robustness under severe class imbalance.

\subsection{Downstream Tasks}
We assess the generalization capability of the pretrained model across two representative downstream tasks:

\vspace{1mm}
\noindent \textbf{Binary Class Collision Detection:}  
    The model predicts whether a collision event occurs within a given time window.  
    This task evaluates sensitivity to rare and transient patterns.  We compare against a statistical baseline GLM that reflects current practice.
    and is consistent with prior statistical approaches to collision detection using acceleration-based sensor data \cite{becker2019collision}. Although GLM is simple, its input variables are carefully engineered using domain knowledge and years of insurance industry experience; these handcrafted features represent a mature modeling pipeline, making it a realistic benchmark in practice. The dataset comprises 112,312 sequences, including 111,200 non-collision and 1,112 collision samples. To evaluate the adaptability of this homogeneous task under increasingly severe label imbalance, we further assess performance using synthetic negative-to-positive ratios of 10:1 and 100:1. These configurations serve as controlled stress tests that reveal how well the pretrained representations adapt with limited task-specific tuning in extreme imbalance settings.

\vspace{1mm}
\noindent \textbf{Multi Class Point of Impact Classification:}  
The model identifies the physical collision location in 8 classes based on CAN signal trajectories during an accident.  
As a baseline, we employ a 1D CNN, a commonly adopted approach in sensor-based fault and anomaly detection tasks \cite{wen2019time, o2015introduction}. The dataset contains 1,088 labeled distributed across eight classes: \textit{rear (21.7\%), front (18.6\%), front-right (15.3\%), front-left (13.8\%), rear-left (9.6\%), rear-right (7.4\%), left (7.0\%), and right (6.6\%)}.

\subsection{Experimental Results and Analysis} 
We first analyze downstream generalization under the condition of data imbalance. In the 10:1 ratio collision-detection setting (see Table~\ref{tab:binary_metrics}), the pretrained foundation model achieves an F1 score of 76.1\%, that is $\sim$5\% below the GLM baseline (81.1\%). The model retains moderate rare-event sensitivity, indicating that the pretrained representations encode meaningful CAN dynamics that transfer to the binary detection objective.

In the multi-class point-of-impact task (Table~\ref{tab:multiclass_metrics}), the foundation model attains a macro-F1 of 27.0\% and a weighted-F1 of 32.6\%, outperforming the CNN baseline by $\sim$3\% and $\sim$5\%, respectively. Since this task differs substantially from binary collision detection, the improvement demonstrates that the learned representations generalize beyond a single problem formulation. These findings support the broader claim that the pretraining→fine-tuning paradigm established in NLP also applies to properly tokenized, decoded-CAN data.

The extremely imbalanced 100:1 collision detection stress test reveals the current model’s limitations. Under this regime, the foundation model’s F1 score decreases to 54.2\%, trailing the baseline’s 65.3\% by $\sim$11\% (Table~\ref{tab:binary_metrics}). This degradation indicates that while the 50M-parameter encoder captures coarse-grained CAN signal structure, it does not yet model the subtle temporal cues required for ultra-rare event discrimination under standard fine-tuning. 
%
While future iterations will address these specific temporal cues, the current results validate the broader utility of our approach, which prioritizes transferable representations and cross-task generalization through a scalable, unified framework, with downstream benefits such as reduced redundant data preparation and training costs, over disjointed task-specific pipelines, as presented in Figure~\ref{fig:architecture}.

Overall, the evaluation confirms that a single pretrained foundation CAN model can transfer meaningfully across tasks. At the same time, the results reveal that the current encoder under-represents fine-grained dynamics needed for extremely rare-event detection. These limitations point to opportunities for future improvements, including larger-scale pretraining, architectural refinement, or more specialized task-aware fine-tuning approaches.

\section{Ablation Study}
We investigate the role of model capacity in CAN representation learning by scaling the foundation model from \textbf{20M} to \textbf{100M} parameters under identical training conditions. Despite a fivefold increase in size, all models converge at approximately \textbf{10k steps} and reach losses nearly identical to the 50M-parameter model. This indicates that the primary bottleneck is the complexity of the data rather than the capacity of the encoder.

This suggests that decoded CAN data, though information-rich, exhibits relatively low compositional entropy compared to natural language, as many normal driving behaviors produce near-repetitive signal patterns. 
Given the current input window of \textbf{10 seconds (450 tokens)}, model size alone does not yield proportional gains.

These findings imply that transformer-based architectures could benefit from richer temporal structure and higher-resolution tokenization.
Future directions include (1) \textbf{extending the temporal window} to capture long-range dependencies and (2) \textbf{adopting finer-grained bin quantization} to encode subtle signal variations using a larger token inventory. 
Additionally, increasing the sampling rate from \textbf{1 Hz to 5 Hz} may reveal finer temporal dynamics, improving transferability across downstream tasks.

Beyond scaling model capacity, we also highlight the importance of diversifying the pretraining corpus across a broader range of driving behaviors, analogous to multi-domain corpora in language model pretraining, which may promote more robust representations and improved generalization across heterogeneous CAN environments.

\section{Limitation and Discussion}
Although the model generalizes across heterogeneous tasks, the pretraining corpus is limited to nine days of driving data. This duration is sufficient for our primary objective, evaluating the feasibility of a foundation-style generalization framework, but it does not capture longer-term seasonal or environmental variability such as snow, rain, or regional driving conditions. Applications that depend on these contextual factors will require substantially broader and more diverse corpora. Extending pretraining across longer time horizons would reduce seasonality bias and better support realistic cross-context generalization. While the evaluated downstream tasks are heterogeneous in label structure, the study is limited to two event-centric objectives, and broader task coverage is left for future work.

Additionally, pretraining was stopped once MLM loss plateaued; however, scaling law analyses \cite{kaplan2020scaling} indicate that in NLP, such plateaus can appear well before the model’s true capacity is reached, particularly when data diversity is constrained. Longer horizon training paired with more heterogeneous CAN corpora may continue to improve representation quality, especially for rare event tasks. While these extensions fall outside the scope of this study, they represent a natural progression toward deploying foundation CAN models in specialized or production-grade settings.



\section{Conclusion and Future Work}
We presented \textbf{Foundation CAN LM}, a pretrained language model for decoded automotive CAN data. 
By treating CAN signals as a structured language, the model achieves cross-task generalization across heterogeneous objectives such as collision detection and point-of-impact prediction. 
Our unified tokenization framework establishes a transferable representation that bridges task-specific CAN modeling and foundation-style learning. 
These findings suggest that properly decoded CAN data can be linguistically modeled, enabling scalable transformer architectures for automotive intelligence. 
In the future, we plan to extend this work to longer temporal contexts, higher-frequency data, and more diverse driving corpora. 


\bibliographystyle{IEEEtran}
\bibliography{references}


\end{document}